\documentclass[letterpaper]{article} 
\makeatletter
\def\copyright@text{}  
\makeatother
\usepackage{aaai2026}  
\usepackage{times}  
\usepackage{helvet}  
\usepackage{courier}  
\usepackage[hyphens]{url}  
\usepackage{graphicx} 
\usepackage{natbib}  
\usepackage{caption} 
\usepackage{algorithm}
\usepackage{algpseudocode}
\usepackage{amsmath, amssymb}

\usepackage{amsmath}
\usepackage{booktabs}
\usepackage{array}
\usepackage{multirow}
\usepackage{bibentry}

\usepackage{tcolorbox, xcolor}
\usepackage{newfloat}
\usepackage{listings}
\DeclareCaptionStyle{ruled}{labelfont=normalfont,labelsep=colon,strut=off} 
\floatstyle{ruled}
\newfloat{listing}{tb}{lst}{}
\floatname{listing}{Listing}
\begin{document}

\title{APVR: Hour-Level Long Video Understanding with Adaptive Pivot Visual Information Retrieval}
\author{
    Hong Gao$^{1,2,*}$ \quad
    Yiming Bao$^{2,*}$ \quad
    Xuezhen Tu$^{2}$ \quad
    Bin Zhong$^{2}$ \quad
    Linan Yue$^{1}$ \quad
    Minling Zhang$^{1,\dagger}$
}
\affiliations{
    $^{1}$SouthEast University, $^{2}$ZTE Corporation\\
    $^{*}$Equal Contribution, $^{\dagger}$Corresponding Author
}

\maketitle

\begin{abstract}
Current multimodal large language models (MLLMs) struggle with hour-level video understanding, facing significant challenges not only in modeling the substantial information volume of long videos but also in overcoming the memory wall and resource constraints during both training and inference. Although recent training-free approaches have alleviated resource demands by compressing visual features, their reliance on incomplete visual information limits the performance potential. To address these limitations, we propose \textbf{A}daptive \textbf{P}ivot \textbf{V}isual information \textbf{R}etrieval (\textbf{APVR}), a training-free framework that hierarchically retrieves and retains sufficient and important visual information. It breakthroughs the memory wall limitation via two complementary components: Pivot Frame Retrieval employs query expansion and iterative spatio-semantic confidence scoring to identify relevant video frames, and Pivot Token Retrieval performs query-aware attention-driven token selection within up to 1024 pivot frames. This dual granularity approach enables the processing of hour-long videos while maintaining semantic fidelity. Experimental validations on three different baseline MLLMs demonstrate significant performance improvements up to 9.5\%, 4.6\% and 9.7\% on LongVideoBench, VideoMME and MLVU, respectively. APVR achieves state-of-the-art results for both training-free and training-based approaches.
\end{abstract}


\section{Introduction}

The proliferation of long-form video content necessitates robust understanding capabilities that extend beyond the current limitations of multimodal large language models (MLLMs). While existing video-based models \cite{lin2023videollava,zhang2025videollama,bai2025qwen2.5vl,team2025kwai} demonstrate proficiency on short video sequences, their scalability for hour-level video remains challenging. Specifically, they encounter not only the memory wall but also the inherent dilution of semantic information across extensive temporal spans.

The core challenge for long video understanding with Video MLLMs lies in efficiently modeling and extracting sparse but critical semantic information across both temporal and spatial dimensions. Existing research generally falls into three paradigms: (1) uniform frame sampling, (2) sparse key-frame retrieval, and (3) sufficient frames processing while requiring computational tricks.

\begin{figure}[!t]
 \begin{center}
 	\centerline{\includegraphics[width= \linewidth]{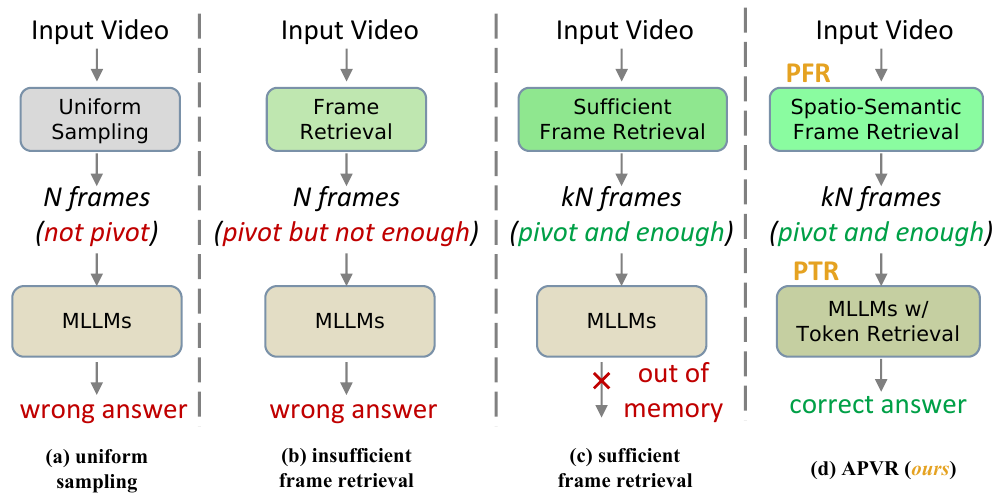}}
\caption{Given a long video, uniform sampling (a) and insufficient frame retrieval (b) yield incorrect answer, while naively increasing pivot frames (c) raises OOM. Our APVR (d) explores joint frame-token co-retrieval, concentrating computational resources on the most relevant information.}
\label{fig:intro}
\end{center}
\end{figure}

The first paradigm, uniform sampling, understands the entire video by evenly selecting frames across the sequence \cite{bai2025qwen2.5vl,zhang2025videollama}. As illustrated in Fig.~\ref{fig:intro}(a), this strategy samples frames that are not relevant and pivot, leading to wrong answer generated by MLLMs. The second line of work attempts to address inefficiency by retrieving only $N$ sparse, query-relevant key frames from long videos \cite{ataallah2024goldfish,tang2025adaptive-aks,guo2025logic,ye2025re,cheng2025scaling-vilamp,luo2025quota}. This key-frame selection greatly reduces the computational load; however, as shown in Fig.~\ref{fig:intro}(b), it neglects the temporal-semantic relationships between events and often fails in scenarios that require comprehensive temporal reasoning or long-range logical understanding. To improve temporal coverage, a third paradigm retrieves a denser set of $kN$ pivot frames (Fig.~\ref{fig:intro}(c)) to capture relevant information. Meanwhile, it quickly encounters the memory wall when fed to MLLMs. Recent solutions leverage training-time trick, such as multimodal sequence parallelism \cite{fu2025vita1.5,liu2024nvila} or feature compression \cite{shu2024videoxl,liu2025videoxlpro}, to handle longer sequences. However, these methods require resource-intensive multi-stage re-training and are tightly coupled to specific MLLM architectures, limiting their adaptability in a rapidly evolving models.

The inherent trade-off between temporal coverage and computational feasibility has hindered existing methods from achieving comprehensive video understanding. To address this core challenge, we propose \textbf{A}daptive \textbf{P}ivot \textbf{V}isual information \textbf{R}etrieval (\textbf{APVR}) framework, a training-free approach that integrates spatio-semantic frame retrieval with token retrieval and compression, as shown in Fig.~\ref{fig:intro}(d). This two-stage retrieval mechanism concentrates computational resources on the most relevant visual information for the input query, thus enabling efficient, scalable and adaptable understanding of hour-long videos without retraining or encountering memory limitations.

Specifically, at the frame level, our \textit{Pivot Frame Retrieval (PFR)} component expands the original query into four types of semantic information: \textit{objects, descriptions, relations,} and \textit{semantics}. This expansion enables more comprehensive frame scoring from spatial and semantic perspectives using complementary visual models: CLIP for semantic similarity and Grounding-DINO for object detection and spatial reasoning. The framework incorporates temporal diffusion mechanisms to maintain temporal coherence and employs adaptive resampling strategies to efficiently and iteratively refine frame scoring and final selection. At the token level, our \textit{Pivot Token Retrieval (PTR)} component extends the concept of semantic importance to fine-grained visual representations. This component leverages query-aware multi-layer attention scoring to identify the most relevant visual tokens to select and the irrelevant visual tokens to compress within selected frames, employing dynamic chunk-wise token selection and head-wise soft voting mechanisms to maintain both computational efficiency and semantic accuracy.

The main contributions of this work are as follows:

\indent $\bullet$ We propose APVR, a novel training-free framework that addresses scalability challenges in long video understanding through dual-granularity information retrieval. Our approach combines frame-level pivot retrieval with token-level adaptive selection to breakthrough the memory wall.

\indent $\bullet$ We develop a mechanism that leverages spatio-semantic confidence scoring and query-aware attention scoring to preserve temporal structure and semantic details, ensuring accurate understanding of complex video narratives while maintaining computational efficiency.

\indent $\bullet$ We demonstrate that intelligent dual-granularity retrieval provides a sustainable alternative to parameter scaling for long video understanding. Our training-free design ensures plug-and-play integration with existing MLLM architectures while preserving foundational capabilities.

\section{Related Work}

\noindent \textbf{Video MLLMs} \quad Recent years have witnessed significant advances in multimodal large language models (MLLMs) for visual perception and understanding. As a critical modality in visual data analysis, video has also been integrated into MLLM frameworks. Notable efforts include Qwen2.5-VL \cite{bai2025qwen2}, InternVideo2.5 \cite{wang2025internvideo2}, VideoLlama3 \cite{zhang2025videollama}, and VideoChat-Flash \cite{li2024videochat}, which explore video understanding through a unified architecture consisting of three key components: (a) a video encoder for spatio-temporal feature extraction, (b) a projector for aligning visual features with linguistic embeddings, and (c) a video-aware LLM for multimodal reasoning. Due to the inherent complexity of video content (e.g., multi-object interactions, long-range temporal dependencies), existing MLLMs often fail to extract semantically salient information for video-centric downstream tasks requiring precise spatiotemporal reasoning, such as video question answering and spatiotemporal video grounding. Some works\cite{bai2025qwen2,wei2025videorope} exploit to aggregate timestamps in the RoPE\cite{su2024roformer}. However, these methods exhibit significant performance degradation when extrapolating to videos exceeding their pre-training duration. Moreover, current architectures also exhibit limitations of inefficient cross-modal alignment between visual information and linguistic queries, particularly in long-form videos exceeding 5 minutes \cite{li2024videomamba}. 

\noindent \textbf{Long Video Understanding} \quad Recently advances in long video understanding can be broadly categorized into training-based and training-free paradigms. In the training-based paradigm, MLLMs are directly trained using large-scale long video datasets. The whole training process is usually divided into several stages to stably improve different components of MLLM, as well as to improve the ability to model different types of input visual data \cite{li2024videochat,zhang2025videollama}. In order to process extra-long videos, current methods explore two different strategies: sequence parallelism \cite{chen2024longvila,longva,shen2025longvita} or visual feature compression \cite{song2024moviechat,shen2024longvu,liu2025videoxlpro}. For instance, LongVILA \cite{chen2024longvila} adopts a five-stage training pipeline with a novel multi-modal sequence parallelism to efficiently learn from long videos. Video-XL-Pro \cite{liu2025videoxlpro} builds a learnable module to generate compact and comprehensive tokens for long video understanding. In the training-free paradigm, long video understanding task is organized as an agent-like process \cite{liu2025videomind,wang2024videoagent}, where MLLMs serve as reasoning engine within the system. These training-free methods mainly focus on retrieving the most important information in video frames \cite{tang2025adaptive,park2024too,ye2025re,guo2025logic,cheng2025scaling-vilamp} or visual tokens \cite{wang2025adaretake,luo2025quota}. For example, AKS \cite{tang2025adaptive} adopts a plug-and-play module for keyframe selection and information pre-filtering for video-based MLLMs. QuoTA \cite{luo2025quota} adopts an ante-hoc training-free module for query-aware visual token assignment and important information retrieval. Despite these advancements, few methods simultaneously explore frame and token optimization, which is addressed in this work through adaptive pivot information retrieval.

\section{Method}
\begin{figure*}[ht]
 \begin{center}
 	\centerline{\includegraphics[width= \linewidth]{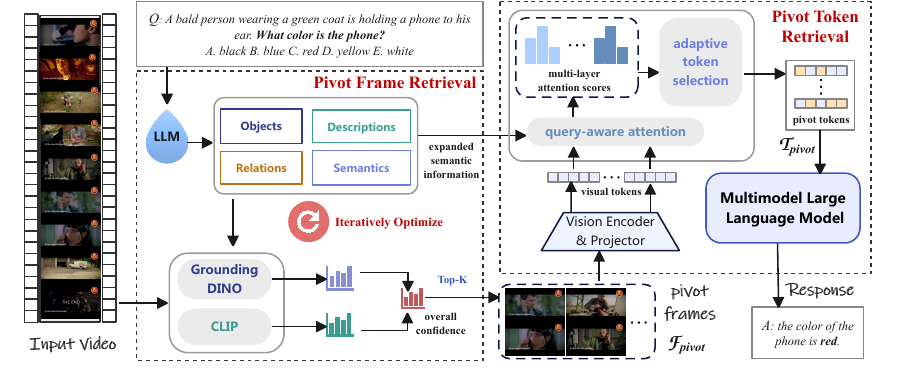}}
\caption{The overall framework of our proposed APVR. We integrate two plug-and-play components, Pivot Frame Retrieval and Pivot Token Retrieval, into MLLMs to improve the performance of long video understanding. APVR is training-free and provides a alternative to parameter scaling. With frame-level and token-level adaptive selection, it can accurately understand complex videos with computational efficiency.}
\label{fig:framework}
\end{center}
\end{figure*}

\subsection{Overview}
The framework of the proposed APVR is illustrated in Fig.~\ref{fig:framework}, which is a training-free system. Given an input video $\mathcal{F} =\{ f_{t} \}_{t=1}^{N}$ with $N$ frames sampled under specific $fps$ and query $Q$, the system aims to retrieve the most important visual information for accurate response. To achieve this, we build two main components in APVR: \textit{Pivot Frame Retrieval} (PFR) and \textit{Pivot Token Retrieval} (PTR). In PFR, we search pivot frames $\mathcal{F}_{pivot}$ with high confidence scores using expanded queries and basic visual models. In PTR, we explore an attention-driven token selection in MLLMs to retrieve pivot tokens $\mathcal{T}_{pivot}$ that are most relevant with expanded queries.

\subsection{Pivot Frame Retrieval}
The progress of PFR is developed in an iterative and efficient manner, as elaborated in Alg. \ref{alg:PFR}.

\subsubsection{Semantic Information Expansion}
First, We expand the given query $Q$ to four types of semantic information as follows:

\noindent \textbf{\textit{Objects}}: The key objects $Obj_k$ or cue objects $Obj_c$ that are detectable by visual models such as Grounding-DINO (e.g., \textit{``bold person'',``phone'',``blue background''}). All objects are aggregated as $Obj = \{Obj_k,Obj_c\}$.

\noindent \textbf{\textit{Descriptions}}: The entities or hypernym concepts of objects based on knowledge graph augmentation, denoted as $Des$. (e.g. \textit{``pearl headband: decorative accessory worn on head; jewelry''})

\noindent \textbf{\textit{Relations}}: The triplet relationship $Rel \subseteq Obj \times \mathcal{R}  \times Obj $  to enhance the confidence of frames with logical links between objects. In each triplet $(o_i,r,o_j)$, the type of relation $r \in \mathcal{R}$ is exactly one of the following: (1) \textbf{spatial} which means that $o_j$ and $o_j$ appear in the same frame, such as \textit{``a bold man holding a red phone''}. (2) \textbf{time} which means that $o_j$ and $o_j$  appear in different frames in orders such as \textit{``A woman walks through, after a cat follows.''} (3) \textbf{attribute} which means that $o_i$ appears with an attribute describable by $o_j$ such as \textit{``A woman in a white skirt''}. (4) \textbf{causal} which means that $o_j$ and $o_j$ appear in orders with a cause-effect manner such as \textit{``A woman opens the door, the cat is scared and runs out.''}

\noindent \textbf{\textit{Semantics}}: The semantics information of query and options based on knowledge graph, denoted as $Sem$ (e.g., \textit{`leash often appears with dog''}).

The prompt template for LLM-based query expansion is provided in Appendix.

\begin{algorithm}[t] 
\caption{Pivot Frame Retrieval (PFR)}
\begin{algorithmic}[1]
\Require video frames $\mathcal{F}$, query $Q$, iterations $P$, stride $\nabla$
\Ensure Pivot frames $\mathcal{F}_{pivot} = \{f_t\}_{t=1}^K$
\State \textbf{Initialize:} 
\State $\mathcal{S} \gets \mathbf{0}^{N}$ \Comment{Confidence scores}
\State $\mathcal{V} \gets \emptyset$ \Comment{Visited frames set}
\State $\{Obj,Des,Rel,Sem\} \gets \operatorname{LLM}(Q)$ \Comment{Query Expansion}
\For{$p = 1$ \textbf{to} $P$}
    \State $\nabla_p \gets \max\left(1, \frac{\nabla}{p}\right)$
    \State $\mathcal{F}_{samp}^p \gets \operatorname{AdaptiveResample}(\mathcal{F}, \mathcal{S}, \mathcal{V}, \nabla_p)$
    \For{$f_t \in \mathcal{F}_{samp}^p$}
        \State $s^{CLIP}_t \gets \Pi_{CLIP}(f_t,Q,Des,Sem)$
        \State $s^{GD}_t \gets \Pi_{GD}(f_t,Obj,Rel)$
        \State $\mathcal{S}_t \gets (1-\lambda) s^{CLIP}_t + \lambda s^{GD}_t$
        \State $\mathcal{S}_{[t-w,t+w]} \gets \max\left( \mathcal{S}_{[t-w,t+w]}, \frac{\mathcal{S}_t}{1+|i-t|} \right)$
    \EndFor
    \State $\mathcal{V} \gets \mathcal{V} \cup \mathcal{F}_{samp}^p$
\EndFor
\State \Return $\mathcal{F}_{pivot} \gets \operatorname{topk}(\mathcal{S} \odot \mathcal{F},K)$
\end{algorithmic}
\label{alg:PFR}
\end{algorithm}

\subsubsection{Spatio-Semantic Confidence Scoring}

Given a video $\mathcal{F}$, and query $Q$ with expanded semantic information $\{Obj,Des,Rel,Sem\}$, instead of exhaustively processing all $N$ frames, we uniformly sample few frames with initial stride $\nabla$ in the first iteration and adaptively resample frames for the subsequent iteration based on the online updated frame confidence. To ensure progressively finer sampling, we gradually decrease the stride by $\nabla^p =\mathrm{max}(1, \frac{\nabla}{p}$). The frames sampled in iteration $p$ are denoted as $\mathcal{F}_{samp}^p$.

Our algorithm exploits two basic visual models, i.e., Grounding-DINO and CLIP, to comprehensively score $\mathcal{F}_{samp}^p$ as follows:

\noindent \textbf{\textit{CLIP-based Similarity Scoring}}: We leverage the CLIP model to compute cross-modal similarity between text and visual features as the semantic confidence score based on expanded query and the visual feature by $s_t^{CLIP} = \Pi_{CLIP}(f_t,Q,Des,Sem)$. The details of the function $\Pi_{CLIP}$ are provided in Appendix.

\noindent \textbf{\textit{Grounding-DINO Object Detection}}: We leverage the Grounding-DINO model to compute the spatial confidence score based on the query-aware objects long with their relations by $s_t^{GD} = \Pi_{GD}(f_t,Obj,Rel)$. The details of the function $\Pi_{GD}$ are provided in Appendix.

The final confidence score of this iteration is computed as a weighted sum of the CLIP and Grounding-DINO scores:
\begin{equation}
    \mathcal{S}_t = (1-\lambda) \cdot s_t^{CLIP} + \lambda \cdot s_t^{GD}.
\end{equation}

The frames scored in the current iteration are put into the visited frame set $\mathcal{V}$. 

\noindent \textbf{\textit{Temporal Diffusion}}: To account for the temporal continuity of video content and improve frame resampling, we introduce a temporal score diffusion mechanism that propagates confidence scores across neighboring frames as:
\begin{equation}
    \mathcal{S}_i = \mathrm{max}(\mathcal{S}_i,\frac{\mathcal{S}_t}{1+|i-t|}), i \in [t-w,t+w].
\end{equation}

This diffusion process helps capture temporally coherent segments relevant to the query. 

\subsubsection{Adaptive Resampling}

Based on the updated $\mathcal{S}$, we further design an adaptive and hybrid candidate sampling strategy for subsequent iterations that combines two types of set:

\noindent \textbf{\textit{High-confidence Set}}: We select frames with higher scores from all unvisited frame set $\mathcal{U} = \{f_t \mid f_t \notin \mathcal{V}\}$ based on the distribution as:
\begin{equation}
    \mathcal{H}_s = \mathrm{TopK}(\mathcal{S} \odot \mathcal{U}, \frac{N}{2\nabla_p}).
\end{equation}

\noindent \textbf{\textit{Uncertainty Set}}: We calculate Shannon entropy based on the score distribution as:
\begin{equation}
    \mathcal{E}_\gamma (s_i) = -\sum_{j=i-\gamma}^{i+\gamma}e_j \cdot \text{log}(e_j), e_j = \frac{s_j}{\sum_{k=j-\gamma}^{j+\gamma} s_k},
\end{equation}
then the frames with high entropy are identified by:
\begin{equation}
    \mathcal{H}_e = \{f_t \mid \mathcal{E}_\gamma(s_t) > \mu+0.5\sigma\},
\end{equation}
where $\mu$ and $\sigma$ are the mean and standard deviation of the entropy in $\mathcal{C}$.

The union candidate set for this iteration sampling is $\mathcal{C} = \mathcal{H}_s \cup \mathcal{H}_e$. Assume that there are $N_\mathcal{C}$ frames in $\mathcal{C}$. The frames $\mathcal{F}_{samp}^p$ are sampled in the way as:

\begin{align}
    \mathcal{F}_{samp}^p = \mathcal{C}_{\text{multi}} \cup \mathcal{C}_{\text{rand}}, \\
     \mathcal{C}_{\text{multi}} \sim \mathrm{Multinomial}(S \odot \mathcal{C},(1-\alpha)N_\mathcal{C}), \\
     \mathcal{C}_{\text{rand}} \sim \mathrm{Unif}(\mathrm{Perm}(S \odot \mathcal{C},\alpha N_\mathcal{C})),    
\end{align}
where $\alpha$ balances the multinomial score-based importance sampling $\mathcal{C}_{\text{multi}}$ and the diverse random sampling $\mathcal{C}_{\text{rand}}$ to capture the full semantic content of the video and also to avoid falling into local optimal.

After $P$ iterations, the $K$ pivot frames are retrieved by $\mathcal{F}_{pivot} = \mathrm{TopK}(\mathcal{S} \odot \mathcal{F},K)$ and fed into the next module.

\begin{table*}[ht]
\begin{center}
\begin{tabular}{lccccccc}
\toprule
\textbf{Methods} & \textbf{Params} & \textbf{Training} & \textbf{LVB} &\multicolumn{2}{c}{\textbf{VideoMME}(w/o sub.)}&\textbf{MLVU}\\
 &&\textbf{-free}&val&Long&Overall&dev\\
\midrule
\textit{\textbf{Proprietary Methods}}\\
GPT4-V\cite{openai2023gpt4v}&-&No&59.1&53.5&59.9&- \\
GPT4-o\cite{hurst2024gpt4o}&-&No&66.7&65.3&71.9&64.6 \\
Gemini-1.5-Pro\cite{team2024gemini}&-&No&64.0&67.4&75.0&- \\
\midrule
\textit{\textbf{Open-Sourse Methods}}\\
VideoLLaVA\cite{lin2023videollava}&7B&No&58.2&36.2&39.9&67.0 \\
VITA-1.5\cite{fu2025vita1.5}&7B&No&-&47.1&56.1&- \\
mPLUG-Owl3\cite{ye2024mplug}&7B&No&52.1&50.1&59.3&63.7 \\
Oryx-1.5\cite{liu2024oryx}&7B&No&56.3&-&58.3&67.5 \\
QuoTA\cite{luo2025quota}&7B&Yes&59.0&55.7&65.9&71.9\\
Video-XL\cite{shu2024videoxl}&7B&No&49.5&-&55.5&64.9 \\
ViLaMP\cite{cheng2025scaling-vilamp}&7B&No&57.8&-&67.5&72.6 \\
AdaReTaKe\cite{wang2025adaretake}&7B&Yes&62.6&58.3&67.7&75.0\\
AKS\cite{tang2025adaptive-aks}&7B&Yes&62.7&-&65.3&-\\
NVILA\cite{liu2024nvila}&8B&No&57.7&54.8&64.2&70.1 \\
ByteVideoLLM\cite{wang2024dynamic-bytevideollm}&14B&No&-&56.4&64.6&65.0 \\
\midrule
Qwen2-VL\cite{wang2024qwen2}&7B&No&55.6&53.8&63.3&66.9 \\
Qwen2-VL w/ APVR(\textit{ours})&7B&Yes&60.9(+9.5\%)&55.1(+2.4\%)&65.2(+3.0\%)&73.4(+9.7\%) \\
\midrule
Qwen2.5-VL\cite{bai2025qwen2.5vl}&7B&No&59.5&55.6&65.4&70.2 \\
Qwen2.5-VL w/ APVR(\textit{ours})&7B&Yes&\textbf{64.9}(+9.1\%)&\textbf{59.1}(+6.3\%)&\textbf{68.4}(+4.6\%)&\underline{76.1}(+8.4\%) \\
\midrule
VideoLLaMA3\cite{zhang2025videollama}&7B&No&59.8&54.9&66.2&73.0 \\
VideoLLaMA3 w/ APVR(\textit{ours})&7B&Yes&\underline{63.5}(+6.2\%)&\underline{58.7}(+6.9\%)&\underline{68.1}(+2.9\%)&\textbf{77.2}(+5.5\%) \\
\bottomrule
\end{tabular}
\caption{Video understanding accuracy(\%) on LongVideoBench (LVB), VideoMME and MLVU. We categorize the methods based on whether they are training-free. The propsed APVR achieved state-of-the-art results on all the three benchmarks and improves the performance based on three different baseline models. Bold denotes the best result while underline the second best. The relative improvement compared to the baseline models are also calculated and listed in the parenthesis.}
\label{table1}
\end{center}
\end{table*}

\begin{figure*}[!ht]
 \begin{center}
 	\centerline{\includegraphics[width= \linewidth]{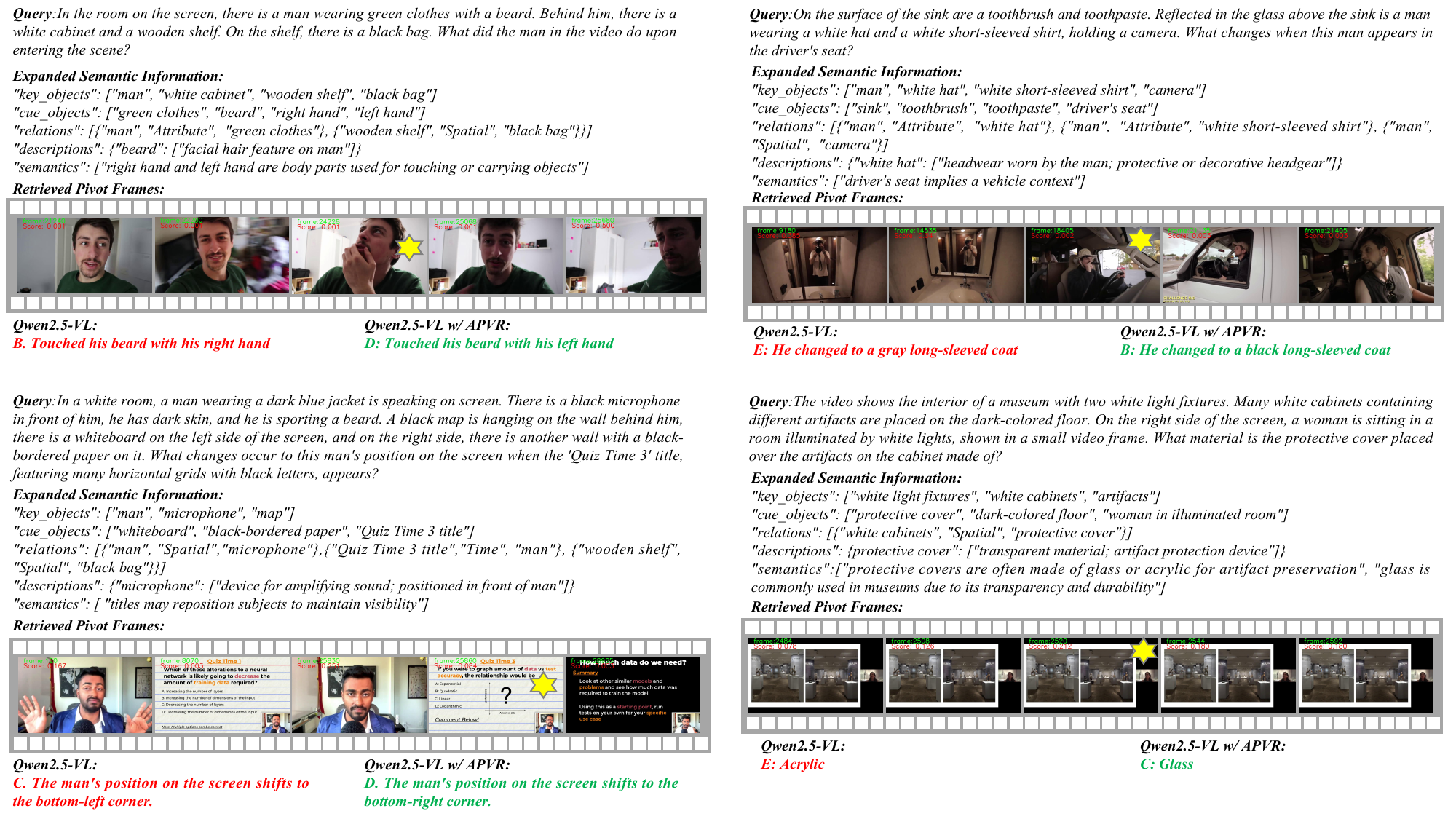}}
\caption{Qualitative Comparison of APVR with the baseline MLLM. The expanded query is significant complementary for pivot information retrieval. The number and score of the pivot frame is drawn in the left-top with green and red text, respectively. Yellow stars indicate the key frame for correct response. Green answer represents correct while red one represents wrong.}
\label{fig:comparison_qualitatively}
\end{center}
\end{figure*}

\subsection{Pivot Token Retrieval}

Pivot Token Retrieval extends the concept of semantic importance from frame-level to token-level granularity. In this module, we adaptively retrieve the pivot visual token within MLLM to maintain computational efficiency while preserving critical semantic information.

First, the retrieved pivot frames $\mathcal{F}_{pivot}$ are processed by a Vision Encoder $\mathcal{G}(\cdot)$ and a Projector $\mathcal{P}(\cdot)$ to generate dense visual token representations: $\mathcal{T}_{vis} = \mathcal{P}(\mathcal{G}(\mathcal{F}_{pivot})) = \{\mathcal{T}_t\}_{t=1}^{K}$. Our objective is to retrieve the pivot token $\mathcal{T}_{pivot}$. To this end, we introduce two components: query-aware multi-layer attention scoring and adaptive token selection.

\subsubsection{Query-aware Multi-Layer Attention Scoring}

We leverage the expanded semantic information from the LLM-augmented query to guide token selection across multiple transformer layers. First, the attention scores for layer $l$ are computed as:
\begin{equation}
    A_{cross}^l = \mathrm{softmax}(\frac{q_{text}^l \cdot (k_{vis}^l)^T}{\sqrt{d}}),
\end{equation}
where $A_{cross}^l \in R^{h\times d\times d_q\times d_v}$, $q_{text}^l$ denotes the query states derived from $\{Q,Des,Sem\}$ and $k_{vis}^l$ denotes the key state derived from the visual tokens $\mathcal{T}_{vis}$. Then the token-level attention scores of each layer are aggregated at the query dimension by:
\begin{equation}
    a^l = \sum_{d_q} A_{cross}^l.
\end{equation}

This formulation captures how much attention each key token receives across all query positions, providing a comprehensive measure of its importance for the current layer. The superscript $l$ will be omitted for concision.

\subsubsection{Adaptive Token Selection}

The simplest token selection strategy is to directly retain the Top-K tokens while evicting others. However, this approach ignores the continuity of visual tokens and may reduce accuracy. Another factor that matters is that attention scores across heads may exist disparities, which further result in independence between heads. Thus, considering the continuity of visual tokens as well as the disparities across heads, we introduce an adaptive token selection strategy consisting of two components: dynamic chunk-wise selection and head-wise soft voting.

\noindent \textbf{\textit{Dynamic Chunk-wise Selection}}: The long video understanding reaches the memory wall when prefilling the visual token. Thus, we perform token selection for the KV cache to reduce redundancy. Given a key or value cache to select, we first divide it into $W$ chunks at the token dimension. Then, the base selection ratio of the chunk $w$ is computed as:
\begin{equation}
    \eta_w = \frac{\sum_{w} a}{\mathrm{max}(\{\sum_i a \mid i=1,...,W\})}.
\end{equation}

The dynamic ratio of each chunk is calculated by:
\begin{equation}
    \rho_w = \mathrm{min}(1.0, \sqrt\frac{|\{j:a_j>0.01 \cdot \mathrm{max}(a)\}|}{L_w}),
\end{equation}
where $L_w$ is the chunk length. The final selection ratio is determined by:
\begin{equation}
    \gamma_w = \rho_w \cdot  \eta_w.
\end{equation}

\noindent \textbf{\textit{Head-wise Soft Voting}}: Based on the ratio $\gamma_w$, we can select the pivot token in each chunk by:
\begin{equation}
    \mathcal{Z}_w = \mathrm{TopK}(a_w \odot \mathcal{T}_w, \gamma_w L_w).
\end{equation}

This results may appear to be significant disparities in different attention heads. To figure out this, we introduce head-wise voting for final token selection:
\begin{equation}
    \mathcal{Z}_w = \mathrm{TopK}((\sum_{j=1}^h \mathrm{softmax}(a_{w,j})) \odot \mathcal{T}_w, \gamma_w L_w),
\end{equation}
where the softmax function calculates the per-head scores and gives a soft voting via normalization and sum. Finally, we update the KV cache as the selected pivot token in each layer:
\begin{equation}
    \mathcal{T}_{pivot} = \mathrm{Concat}(\{\mathcal{Z}_w\}_{w=1}^W)
\end{equation}.

Finally, the selected pivot tokens will be fed into the MLLM to generate the final response.

\begin{table}[!t]
\begin{center}
\begin{tabular}{l|cc}
\toprule
\specialrule{0em}{1pt}{1pt}
Method& Long (10-60m) & Avg\\
\specialrule{0em}{1pt}{1pt}
\midrule
\specialrule{0em}{1pt}{1pt}
APVR (Qwen2-VL-7B)&51.2&60.9 \\
w/o exp. sem. info.&50.4&59.1 \\
\midrule
\specialrule{0em}{1pt}{1pt}
APVR (Qwen2.5-VL-7B)&58.5&64.5 \\
w/o exp. sem. info.&56.6&63.2 \\
\midrule
\specialrule{0em}{1pt}{1pt}
APVR (VideoLLaMA3-7B)&56.9&63.5 \\
w/o exp. sem. info.&55.5&62.3 \\
\bottomrule
\end{tabular}
\end{center}
\caption{Ablations of expanded semantic information.}
\label{table_esi}
\end{table}

\section{Experiments}

\subsection{Experimental Setup}
\subsubsection{Datasets}
We evaluate the performance of APVR in three popular benchmarks that contains hour-level videos: \textbf{LongVideoBench}\cite{wu2024longvideobench}: A benchmark consisting of 17 categories of referred reasoning questions.  \textbf{VideoMME}\cite{fu2024videomme}: A multi-modal evaluation benchmark focused on fine-grained video understanding tasks. \textbf{MLVU}\cite{zhou2025mlvu}: A large-scale long video benchmark with a large wide of 9 distinct tasks. To highlight the importance of the pivot information retrieval for visual understanding, we do not use subtitles for all benchmarks.


\subsubsection{Implementation Details}
We integrate APVR with three baseline MLLMs, namely, Qwen2-VL-7B \cite{wang2024qwen2}
, Qwen2.5-VL-7B \cite{bai2025qwen2.5vl} and VideoLLaMA3 \cite{zhang2025videollama}. The MLLMs process a prompt involving the question, retrieved pivot video frames, and the multi-choice answer options to generate responses. We first densely extracted the raw frames from video with $fps=2$. Then, the sampled frames are iteratively scored using two basic visual models with $\lambda=0.5$: CLIP \cite{radford2021learning-clip} with ViT-B-16 \cite{dosovitskiy2020image-vit} backbone, and Grounding-DINO \cite{liu2024grounding-dino} with Swin-T \cite{liu2021swinT} backbone, based on the expanded semantic information generated by LLM. All evaluations are conducted on 8 NVIDIA A800 GPUs with 80GB memory, using LMMs-Eval \cite{zhang2024lmmseval} framework. Under the above setup, our APVR enables processing of up to $K=1024$ frames, compared to the baseline 7B model's capacity of $256$ frames under optimal MLLM configurations. Our APVR can understand an hour-level long video and generate correct answers in up to 2 minutes.

\subsection{Comparison to the State-of-the-Art}

We first compare the accuracy of video question answering between APVR with recent training-based or training-free methods. All results are summarized in Table \ref{table1}. Upon three different baseline MLLMs, APVR improves overall accuracy on all three widely used benchmarks. Notably, APVR based on Qwen2.5-VL reaches 64.9\% accuracy in LongVideoBench, which is higher than the powerful proprietary models such as GPT4-V and Gemini-1.5-Pro. The overall performance of APVR also surpasses other training-based and training-free methods with larger parameter scales such as 8B and 14B. The complete verification on the benchmarks confirms the capability of APVR to preserve critical spatio-temporal information in long video sequences.

\begin{figure}[t]
 \begin{center}
 	\centerline{\includegraphics[width= 0.9\linewidth]{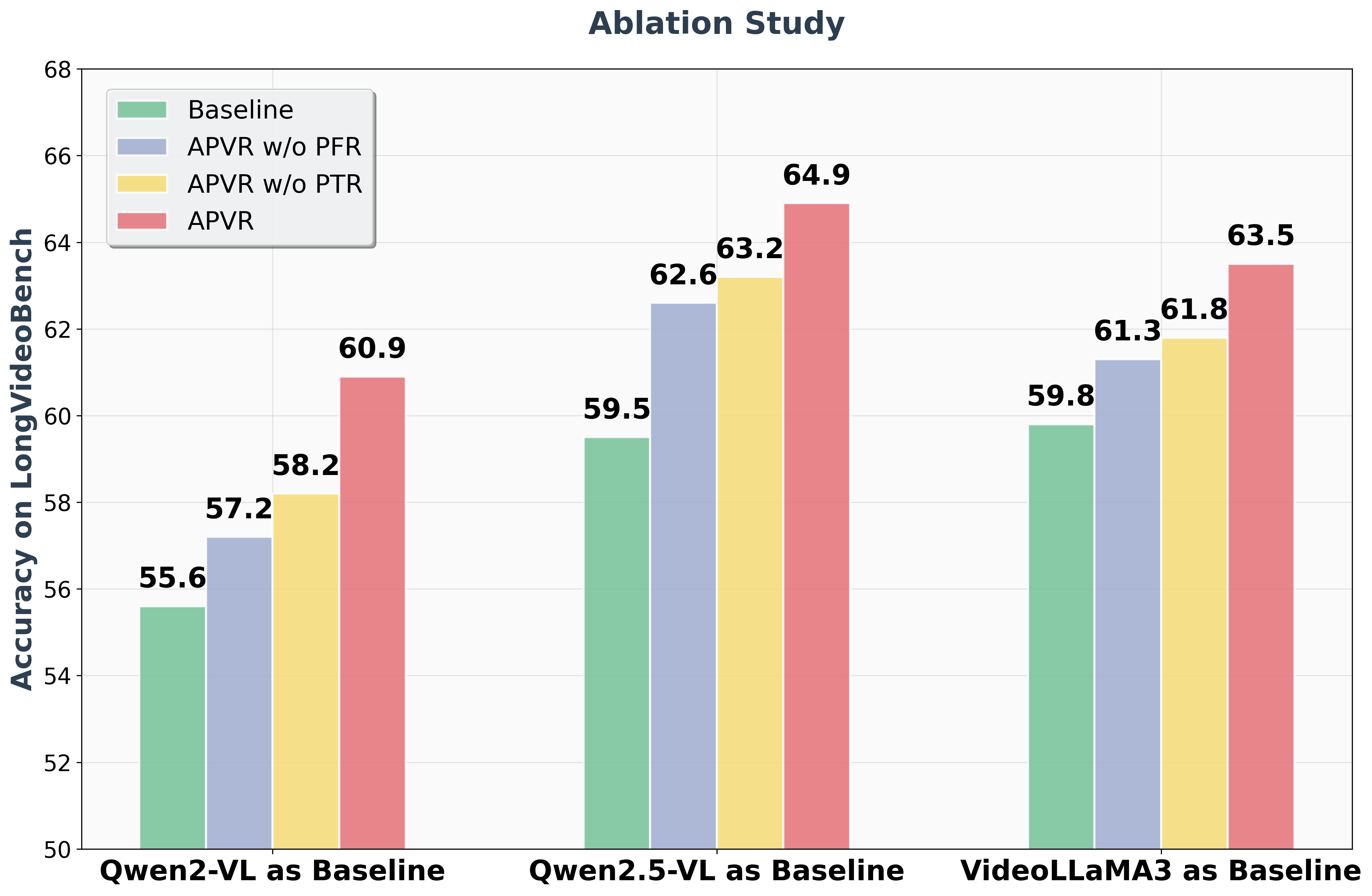}}
\caption{The result of ablation study on LongVideoBench for the two designed components: PFR and PTR.}
\label{fig:ablation_components}
\end{center}
\end{figure}

\subsection{Analysis on Information Retrieval}

Fig.~\ref{fig:comparison_qualitatively} presents representative qualitative results on hour-level video understanding, comparing our APVR framework with the baseline Qwen2.5-VL across four challenging samples. The results clearly demonstrate the advantages of APVR in both the semantic expansion and the subsequent retrieval process.

Firstly, we observe that the pivot frames selected by APVR are highly correlated with both the query and the expanded semantic information. APVR distills the most informative visual information within sufficient pivot frames. As a result, APVR empowers MLLM to efficiently comprehend long-form video content, providing accurate and detailed answers even in scenarios where baselines fail or generate hallucinations. Secondly, the semantic information expansion module in APVR significantly enriches the original query, providing the downstream retrieval process with comprehensive object, attribute, and relational cues that are otherwise missing when using only the raw query. As shown by ablation study results in Table~\ref{table_esi}, on LongVideoBench the performances decrease without expanded semantic information no matter the baseline MLLM.

\begin{table}[!t]
\begin{center}
\begin{tabular}{ccc|ccc}
\toprule
\specialrule{0em}{1pt}{1pt}
$K$ & Long & Average & $\lambda$ & Long & Average\\
\specialrule{0em}{1pt}{1pt}
\midrule
\specialrule{0em}{1pt}{1pt}
1024&58.3&64.9 &0.1&57.4&64.2\\
256&56.4&63.1&0.2&57.8&64.5 \\
128&53.0&61.1 &0.5&58.3&64.9\\
64&49.8&58.6 &0.8&58.0&64.8\\
32&45.4&54.3 &0.9&58.9&64.5\\
\bottomrule
\end{tabular}
\end{center}
\caption{Ablations of selected frames $K$ and weight $\lambda$.}
\label{table_topk_and_w}
\end{table}

\subsection{Ablation Studies on PFR and PTR}

To systematically validate the effectiveness of the proposed dual-granularity retrieval strategy, we conduct ablation studies for the two core components of APVR: Pivot Frame Retrieval (PFR) and Pivot Token Retrieval (PTR), on LongVideoBench. The results, shown in Fig.~\ref{fig:ablation_components}, consistently demonstrate that removing either PFR or PTR leads to a notable decline in accuracy across all baseline MLLMs. Without PFR, the system loses its ability to precisely localize query-relevant frames, resulting in performance degradation due to the inclusion of irrelevant content (Fig.~\ref{fig:intro}(a)). Without PTR, the model is less able to compress visual tokens and focus on the most informative visual details, leading to processing maximum to only 256 frames and reduced answer quality (Fig.~\ref{fig:intro}(b) and (c)). These findings strongly support the necessity and rationality of our dual-granularity pivot information retrieval design.

\subsection{Ablation Studies on Hyper-Parameters}

We further conduct a comprehensive analysis of key hyper-parameters that influence the performance of APVR by ablation studies on LongVideoBench. First, we examine the impact of the number of retrieved pivot frames $K$ and the frame scoring weights $\lambda$. As shown in Table \ref{table_topk_and_w}, increasing $K$ generally improves performance, especially on long videos, by reducing the risk of missing relevant segments. For $\lambda$, we observe that optimal results are achieved by striking a balance between spatial and semantic scoring, confirming the complementarity of these two perspectives.

Second, we analyze the effect of  the number of frame retrieval iterations $p$ and the initial sampling stride $\nabla$. As shown in Table \ref{table_s_and_p}, the best performance is observed when $P=3$ and $\nabla=4$, indicating that a moderate number of adaptive iterations and a balanced initial sampling density allow the system to progressively refine frame selection without causing overfitting or excessive computational cost. 


\begin{table}[!t]
\begin{center}
\begin{tabular}{cc|ccc}
\toprule
\specialrule{0em}{1pt}{1pt}
$P$&$\nabla$ & Long(10-60m) & Medium(1-10m)& Average\\
\specialrule{0em}{1pt}{1pt}
\midrule
\specialrule{0em}{1pt}{1pt}
3&2&56.4&62.1&62.9 \\
3&4&58.3&65.8&64.9 \\
3&8&55.1&64.6&63.1 \\
2&4&56.0&66.0&63.9 \\
4&4&58.0&65.8&64.7 \\
5&4&58.5&64.5&64.5 \\
\bottomrule
\end{tabular}
\end{center}
\caption{Ablations of iteration number $P$ and stride $\nabla$.}
\label{table_s_and_p}
\end{table}

\section{Conclusion}

We propose APVR, a training-free framework that overcomes the memory wall in hour-level video understanding via adaptive, hierarchical visual information retrieval. By integrating Pivot Frame Retrieval and Pivot Token Retrieval, APVR enables efficient and accurate processing of long videos without sacrificing semantic detail. Extensive experiments on standard benchmarks demonstrate that APVR delivers state-of-the-art results among training-free and training-based methods. Its plug-and-play design allows seamless integration with existing MLLM systems, eliminating the need for costly retraining. Our results establish retrieval-based optimization as a practical and scalable solution for long-form video comprehension.

\bibliography{aaai2026}

\section{Appendix}

\subsection{Prompt Template for Query Expansion}

\noindent
The prompt in Fig.~\ref{fig:prompt} was used to interact with the language model for semantic information expansion.

\begin{figure*}
\begin{tcolorbox}[colframe=black!40!white, colback=gray!10!white, sharp corners=south, title=\textbf{Query Expansion Prompt Example}]
Analyze the following video understanding question:

Question: \textit{\textless Question\textgreater}; Options: \textit{\textless Options\textgreater}

\textbf{Step 1}: Key Object Identification

    \quad • Extract 3-5 core objects detectable by computer vision

    \quad • Use Grounding dino compatible noun phrases (e.g., ``person'', ``mic'')
    
    \quad • Format: Key Objects: obj1, obj2, obj3

\textbf{Step 2}: Contextual Cues

    \quad • List 2-4 scene elements that help locate key objects based on options provided
    
    \quad • Use detectable items (avoid abstract concepts)
    
    \quad • Format: Cue Objects: cue1, cue2, cue3

\textbf{Step 3}: Relationship Triplets

    \quad • Relationship types:
    
    \quad \quad • Spatial: Objects must appear in the same frame
    
    \quad \quad • Attribute: Color/size/material descriptions (e.g., ``red clothes'', ``large'')
    
    \quad \quad • Time: Appear in different frames within a few seconds
    
    \quad \quad • Causal: There is a temporal order between the objects
    
    \quad • Format of Relations: (object, relation type, object), relation type should be exactly one of spatial/attribute/time/causal

\textbf{Step 4}: Description Augmentation

    \quad • List 1-3 descriptions based on knowledge graph augmentation for each object
    
    \quad • Entity descriptions (e.g., ``mic is a device for amplifying sound'')
    
    \quad • Hypernym concepts (e.g., ``dog is a kind of animal'')
    
    \quad • Format: Description: (object: des1; des2)

\textbf{Step 5}: Semantics Augmentation

    \quad • List 2-5 Semantics information of query and options based on knowledge graph (e.g., ``leash often appears with dog'')
    
    \quad • Format: Semantics: semantic1; semantic2

\vspace{1em}

\textbf{Output Rules}

\quad 1. One line each for Key Objects/Cue Objects/Relation/Des/Sem starting with exact prefixes

\quad 2. Separate items with comma except for triplets where items are separated by semicolon

\quad 3. Never use markdown or natural language explanations

\quad 4. If you cannot identify any key objects or cue objects from the video provided, please just identify the possible key or cue objects from the question and options provided

\vspace{1em}

\textbf{Below is an example of the procedure:}

Question: For ``When does the person in red clothes appear with the dog?''

Response:

\quad Key Objects: person, dog, red clothes

\quad Cue Objects: grassy area, leash, fence

\quad Rel: (person; attribute; red clothes), (person; spatial; dog)

\quad Des: (red clothes: description1), (dog: description2) 

\quad Sem: semantic1; semantic2

\textbf{Format your response EXACTLY like this in Five lines:}

\quad Key Objects: object1, object2

\quad Cue Objects: object3, object4

\quad Rel: (object1; relation type1; object2), (object3; relation type2; object4), 
...

\quad Des: (object1: description1; description2), (object2: description1; description2), ... 

\quad Sem: semantic1; semantic2, ...
\end{tcolorbox}
\caption{The Prompt Template for Quey Expansion.}
\label{fig:prompt}
\end{figure*}

\subsection{The Details of Spatio-semantic Confidence Scoring}

\subsubsection{Semantic Scoring}

For each of the sampled frames $\mathcal{F}_{samp} = \{f_t\}_{t=1}^{N_s}$, we encode it using the CLIP image encoder $\phi_I(\cdot)$:
\begin{equation}
    v_t = \frac{\phi_I(f_t)}{||\phi_I(f_t)||_2}.
\end{equation}

The query as well as the expanded description and semantics are aggregated and encoded using the CLIP text encoder $\phi_T(\cdot)$:
\begin{equation}
    t_{agg} = \sum_{e}\frac{\phi_T(e)}{||\phi_T(e)||_2}, e \in \{Q,Des,Sem\}.
\end{equation}

The CLIP similarity score for frame $f_t$ is then calculated as:
\begin{equation}
    s_t^{CLIP} = softmax(\tau \cdot v_t \cdot t_{agg}),
\end{equation}
where $\tau$ is a temperature parameter set to $100$ to sharpen the score distribution.

\subsubsection{Spatial Scoring}

To enhance the retrieval precision with key/cue objects along with their relations, we first employ Grounding-DINO to identify extracted objects:
\begin{equation}
    \mathcal{B}_t,\mathcal{L}_t = \Psi_{GD}(f_t,O_k \cup O_c),
\end{equation}
where $\Psi(\cdot)$ is the Grounding-DINO model that returns bounding boxes $\mathcal{B}_t$ and corresponding confidence logits $\mathcal{L}_t$ for frame $f_t$. The score of detection is then computed as:
\begin{equation}
    s_t^{o} = softmax(max(\mathcal{L}_t)).
\end{equation}

To further model the logic relations among objects in the same or different frames to enhance the detection results, we update the score by adding corresponding frame score if there are satisfied relation triplets in this frame as follows:
\begin{equation}
    s_t^{GD} = s_t^o + \sum_r (\omega_r \cdot s_r),
\end{equation}
where $r$ is exactly one of spatial/time/attribute/causal and $\omega_r$ is the corresponding weight to control the contribution of each type of relation.

\end{document}